\documentclass[final]{article}

\usepackage{arxiv}

\usepackage[utf8]{inputenc} 
\usepackage[T1]{fontenc}    
\usepackage{hyperref}       
\usepackage{url}            
\usepackage{booktabs}       
\usepackage{amsfonts}       
\usepackage{nicefrac}       
\usepackage{microtype}      
\usepackage{lipsum}
\usepackage{graphicx}

\usepackage{amsmath,amsfonts,bm}

\def\eqref#1{equation~\ref{#1}}

\def\1{\bm{1}}

\DeclareMathAlphabet{\mathsfit}{\encodingdefault}{\sfdefault}{m}{sl}
\SetMathAlphabet{\mathsfit}{bold}{\encodingdefault}{\sfdefault}{bx}{n}

\newcommand{\E}{\mathbb{E}}

\newcommand{\R}{\mathbb{R}}

\DeclareMathOperator*{\argmax}{arg\,max}
\DeclareMathOperator*{\argmin}{arg\,min}

\usepackage{natbib}
\usepackage{hyperref}
\usepackage{url}
\usepackage{graphicx}
\usepackage{mathtools}

\begin{document}

%

%

\title{Addressing Variance Shrinkage in Variational Autoencoders using Quantile Regression}

\author{
   Haleh Akrami \\
   Signal and Image Processing Institute,\\ 
   Ming Hsieh Department of Electrical and Computer Engineering,\\ 
   University of Southern California,\\
   Los Angeles, CA, USA\\
   \texttt{email: akrami@usc.edu} \\
  \And
   Anand A. Joshi \\
   Signal and Image Processing Institute,\\ 
   Ming Hsieh Department of Electrical and Computer Engineering,\\ 
   University of Southern California,\\
   Los Angeles, CA, USA\\
   \texttt{email: ajoshi@usc.edu}
  \And
   Sergul Aydore \\
   Amazon Web Services,\\
   New York, NY, USA\\
   \texttt{email: sergulaydore@gmail.com}
  \And
   Richard M. Leahy \\
   Signal and Image Processing Institute,\\ 
   Ming Hsieh Department of Electrical and Computer Engineering,\\ 
   University of Southern California,\\
   Los Angeles, CA, USA\\
   \texttt{email: leahy@sipi.usc.edu}
}

\maketitle

\begin{abstract}

Estimation of uncertainty in deep learning models is of vital importance, especially in medical imaging, where reliance on inference without taking into account uncertainty could lead to misdiagnosis.
Recently, the probabilistic Variational AutoEncoder (VAE) has become a popular model for anomaly detection in applications such as lesion detection in medical images. The VAE is a generative graphical model that is used to learn the data distribution from samples and then generate new samples from this distribution. By training on normal samples, the VAE can be used to detect inputs that deviate from this learned distribution. The VAE models the output as a conditionally independent Gaussian characterized by means and variances for each output dimension. VAEs can therefore use reconstruction probability instead of reconstruction error for anomaly detection. Unfortunately, joint optimization of both mean and variance in the VAE leads to the well-known problem of shrinkage or underestimation of variance. We describe an alternative approach that avoids this variance shrinkage problem by using quantile regression. 
Using estimated quantiles to compute mean and variance under the Gaussian assumption, we compute reconstruction probability as a principled approach to outlier or anomaly detection. 
Results on simulated and Fashion MNIST data demonstrate the effectiveness of our approach. We also show how our approach can be used for principled heterogeneous thresholding for lesion detection in brain images.
\end{abstract}
\section{Introduction}
\label{Sec:intro}

Inference based on deep learning methods that does not take into account uncertainty can lead to over-confident predictions, particularly with limited training data \citep{reinhold2020validating}. Quantifying uncertainty is particularly important in critical applications such as clinical diagnosis, where a realistic assessment of uncertainty is essential in determining disease status and appropriate treatment. Here we address the important problem of learning uncertainty in order to perform statistically-informed inference.

Unsupervised learning approaches such as the Variational autoencoder (VAE) \citep{kingma2013auto} and its variants \citep{makhzani2015adversarial} can approximate the underlying distribution of high-dimensional data. VAEs are trained using the variational lower bound of the marginal likelihood of data as the objective function. They can then be used to generate samples from the data distribution, where probabilities at the output are modeled as parametric distributions such as Gaussian or Bernoulli that are conditionally independent across output dimensions \citep{kingma2013auto}. 

VAEs are popular for anomaly detection. Once the distribution of anomaly-free samples is learned, during inference we can compute the reconstruction error between a given image and its reconstruction to identify abnormalities \citep{aggarwal2015outlier}. Decisions on the presence of outliers in the image are often based on empirically chosen thresholds.  \citet{an2015variational} proposed to use reconstruction probability rather than the reconstruction error to detect outliers. This allows a more principled approach to anomaly detection since inference is based on quantitative statistical measures and can include corrections for multiple comparisons.   

Predictive uncertainty can be categorized in two types: aleatoric uncertainty as a result of unknown or unmeasured features and epistemic uncertainty which is often referred as model uncertainty, as it is the uncertainty due to model limitations \citep{skafte2019reliable}. Infinite training data do not reduce the former uncertainty in contrast to the latter. \citet{skafte2019reliable} summarized different methods to estimate uncertainty including Gaussian processes, uncertainty aware neural networks, Bayesian neural networks, and ensemble methods. Another recent approach makes use of collaborative networks \citep{zhou2020estimating}. While we focus here on anomaly 
detection,  estimating uncertainty has a wide range of applications including reinforcement learning, active learning, and Bayesian optimization \citep{szepesvari2010algorithms,huang2010active,frazier2018tutorial,ross2008bayesian,reinhold2020validating}.

Here, we focus on predicting variance use VAEs.  These variances represent an aleatoric uncertainty associated with the  conditional variance of the estimates given the data \citep{reinhold2020validating}. Estimating the variance is more challenging than estimating the mean in generative networks due to the unbounded likelihood \citep{skafte2019reliable}. In the case of VAEs, if the conditional mean network prediction is nearly perfect (zero reconstruction error), then maximizing the log-likelihood pushes the estimated variance towards zero in order to maximize likelihood. This also makes VAEs susceptible to overfitting the training data. These near-zero variance estimates, with the log-likelihood approaching an infinite supremum, do not lead to a good generative model. It has been shown that there is a strong link between this likelihood blow-up and the mode-collapse phenomenon \citep{mattei2018leveraging,reinhold2020validating}. In fact, in this case, the VAE behaves much like a deterministic autoencoder \citep{blaauw2016modeling}. 

VAEs are among the most widely used generative models. However, while the classical formulation of VAEs allows both mean and variance estimates \citep{kingma2013auto}, because of the variance shrinkage problem, most if not all implementations estimate only the mean with the variance assumed constant \citep{skafte2019reliable}.

Here we describe an approach that overcomes the variance shrinkage problem in VAEs using quantile regression (QR) in place of variance estimation. We then demonstrate this new QR-VAE by computing reconstruction probabilities for an anomaly detection task.

\textbf{Related Work:}
 A few recent papers have targeted the variance shrinkage problem. Among these, \citet{detlefsen2019reliable} describe reliable estimation of the variance using Comb-VAE, a locally aware mini-batching framework that includes a scheme for unbiased weight updates for the variance network. In an alternative approach, \citet{stirn2020variational}
suggest treating variance variationally, assuming a Student’s t likelihood for the posterior to prevent optimization instabilities and assume a Gamma prior for the precision parameter of this distribution. 
The resulting
Kullback–Leibler (KL) divergence induces gradients that prevent the
variance from approaching zero \citep{stirn2020variational}.

\textbf{Our Contribution:} To address the variance shrinkage problem, we suggest an alternative and attractively simple solution: assuming the output of the VAE has a Gaussian distribution, we quantify uncertainty in VAE estimates using conditional quantile regression (QR-VAE). The aim of conditional quantile regression \citep{koenker1978regression} is to estimate a quantile of interest. Here we use these quantiles to compute variance, thus sidestepping the shrinkage problem. It has been shown that quantile regression is able to capture aleatoric uncertainty  \citep{tagasovska2019single}. We demonstrate the effectiveness of our method quantitatively and qualitatively on simulated and real-world datasets. Our approach is  computationally efficient and does not add any complication to training or sampling procedures.

\section{Background}
Before introducing our approach, we first define mathematical notations and briefly explain the VAE formulation and the variance shrinkage problem. We also summarize the conditional quantile regression formulation.

\subsection{Variance Shrinkage Problem in Variational Autoencoders}
Let $x_{i} \in \mathbb{R}^D$ be an observed sample of random variable ${X}$ where $i \in \{1, \cdots, N \}$, $D$ is the number of features and $N$ is the number of samples; and let ${z_{i}}$ be an observed sample for latent variable ${Z}$ where $j \in \{1, \cdots, S \}$. Given a sample $x_{i}$ representing an input data, VAE is a probabilistic graphical model that estimates the posterior distribution $p_{\theta}({Z}|{X})$ as well as the model evidence $p_{\theta}({X})$, where $\theta$ are the generative model parameters \citep{kingma2013auto}. The VAE approximates the posterior distribution of ${Z}$ given ${X}$ by a tractable parametric distribution and minimizes the ELBO loss \citep{an2015variational}. It consists of  the encoder network that computes $q_{\phi}(Z|X)$,
and the decoder network that computes $p_{\theta}(X|Z)$  \citep{wingate2013automated}, where ${\phi}$ and ${\theta}$ are model parameters. The marginal likelihood of an individual data point can be rewritten as follows:
\begin{align}
\begin{split}
\log p_{\theta}(x_{i}) = & D_{KL}(q_{\phi}(Z|x_{i}),p_{\theta}(Z|x_{i}))\\
 & + L(\theta,\phi;x_{i}),
\end{split}
\end{align}
where
\begin{align}
\begin{split}
    L(\theta,\phi;x_{i})=& \E_{q_{\phi}(Z|x_{i})}[\log(p_{\theta}(x_{i}|Z))]\\ &-D_{KL}(q_{\phi}(Z|x_{i})||p_{\theta}(Z)).
\end{split}
\label{eqn:overall_loss}
\end{align}

The first term (log-likelihood) in equation \ref{eqn:overall_loss} can be interpreted as the  \emph{reconstruction loss} and the
second term (KL divergence) as the \emph{regularizer}. The total loss over all samples can be written as:
\begin{align}
L({\theta},{\phi},X) &= L_{REC}+L_{KL}
\end{align}
where $L_{REC} \coloneqq \mathbb{E}_{q_{\phi}({Z|X})}[\log(p_{\theta}({X}|{Z}))]$
and $ L_{KL} \coloneqq  D_{KL}(q_{\phi}({Z|X})||p_{\theta}({Z}))$.

Assuming the posterior distribution is Gaussian and using 1-sample approximation \citep{skafte2019reliable}, the likelihood term  simplifies to:
\begin{align}
L_{REC}=\sum_i\frac{-1}{2}log(\sigma_{\theta}^{2}(x_{i}))-\frac{(x_{i}-\mu_{\theta}(z_{i}))^{2}}{2\sigma_{\theta}^{2}(z_{i})}
\end{align}
where $p(Z)=  \mathcal{N}(0,\,I)$, $p_{\theta}(X|Z)=  \mathcal{N}(X|\mu_{\theta}(Z),\,\sigma_{\theta}(Z))$, and $q_{\phi}(Z|X)=  \mathcal{N}(Z|\mu_{\phi}(X),\,\sigma_{\phi}(X))$. 

Optimizing VAEs with a Gaussian posterior is difficult \citep{skafte2019reliable}. If the model has sufficient capacity that there exists $(\phi,\theta)$ for which $\mu_{\theta}(z)$ provides a sufficiently good reconstruction, then the second term pushes the variance to zero before the term $\frac{-1}{2}log(\sigma_{\phi}^{2}(x_{i})))$ catches up \citep{blaauw2016modeling,skafte2019reliable}. 

One good example of this behavior is in speech processing applications \citep{blaauw2016modeling}. The input is a spectral envelope which is a relatively smooth 1D curve.  Representing this as a 2D image produces highly structured and simple training images. As a result, the model  quickly learns how to accurately reconstruct the input. Consequently, reconstruction errors are small and the estimated variance becomes vanishingly small. 
Another example is 2D reconstruction of MRI images where the images from neighbouring 2D slices are highly correlated leading again to variance shrinkage \citep{volokitin2020modelling} . 
To overcome this problem, variance estimation networks are avoided by using a Bernoulli distribution or  the variance is simply set to a constant value \citep{skafte2019reliable}. 

\subsection{Conditional Quantile Regression }
In contrast to classical parameter estimation where the goal is to estimate the conditional mean of the response variable given the feature variable, the goal of quantile regression is to estimate conditional quantiles based on the data  \citep{yu2001bayesian,koenker2001quantile}. 
The most common application of quantile regression models is in the cases in which parametric likelihood cannot be specified \citep{yu2001bayesian}.

Quantile regression can be used to estimate the conditional median (0.5 quantile) or other quantiles of the response variable conditioned on the feature variable. The $\alpha$-th conditional quantile function is defined as $q_{\alpha}(x) \coloneqq \inf\{y\in\R:F(y|X=x)\geq \alpha\}$ where $F=P(Y \leq y)$ is strictly monotonic cumulative distribution function. Similar to classical regression analysis which estimates the conditional mean, the $\alpha$-th quantile regression $(0<\alpha<1)$ seeks a solution to the following minimization problem for input $x$ and output $y$ \citep{koenker1978regression,yu2001bayesian}:
\begin{align}
\argmin \limits_{\theta}\sum_{i}\rho_{\alpha}(y_{i}-f_{\theta}(x_{i}))
\label{q_loss}
\end{align}

where $x_i$ are the inputs, $y_i$ are the responses,  $\rho_{\alpha}$ is the \emph{check function} or \emph{pinball loss} \citep{koenker1978regression} and $f$ is the model paramaterized by $\theta$. The goal is to estimate the parameter $\theta$ of the model $f$. The  \emph{pinball loss} is defined as:
\begin{align}
\rho_{\alpha}(y,\hat y):=\begin{cases}
            \alpha(y-\hat y)   &\text{if $(y-\hat y) > 0$ }  \\
            (1-\alpha)(y-\hat y)   &\text{otherwise.}
        \end{cases}
\end{align}

Due to its simplicity and generality, quantile regression is widely applicable in
classical regression and machine learning to obtain a conditional
prediction interval \citep{rodrigues2020beyond}.
It can be shown that minimization of the loss function in equation \ref{q_loss}  is  equivalent to maximization of the likelihood function formed by combining independently distributed asymmetric Laplace densities \citep{yu2001bayesian}:
\begin{align*}
\argmax \limits_{\theta} L(\theta)=\frac{\alpha(1-\alpha)}{\sigma} \exp\left\{\frac{-\sum_i \rho_{\alpha}(y_{i}-f_{\theta}(x_{i}))}{\sigma}\right\}
\end{align*}
where $\sigma$ is the scale parameter.

\section{Proposed Approach: Uncertainty Estimation for Autoencoders with Quantile Regression (QR-VAE)}

Instead of estimating the conditional mean and conditional variance directly at each pixel (or feature), the outputs  of our QR-VAE are multiple quantiles of the output distributions at each pixel. This is done by replacing the Gaussian likelihood term in the VAE loss function with the quantile loss (check or pinball loss). For the QR-VAE, if we assume a Gaussian output, then only two quantiles are needed to fully characterize the Gaussian distribution. Specifically, we estimate the median and $0.15$-th quantile, which corresponds to one standard deviation from the mean. Our QR-VAE ouputs, $Q_{L}$ (low quantile) and $Q_{H}$ (high quantile), are then used to calculate the mean and the variance. To find these conditional quantiles, fitting is achieved by
minimization of the pinball loss for each quantile. The resulting reconstruction loss for the proposed model can be calculated as:
\begin{align*}
L_{REC}=\sum_{i}\rho_{L}(x_{i}-f_{\theta_L}(x_{i}))+\sum\rho_{H}(x_{i}-f_{\theta_H}(x_{i}))
\end{align*}
where $\theta_L$ and $\theta_H$ are the parameters of the models corresponding to the quantiles $Q_L$ and $Q_H$, respectively.
These quantiles are estimated for each output pixel or dimension.

We prevent quantile crossing \citep{he1997quantile} by limiting the flexibility  of independent quantile regression since both quantiles are estimated simultaneously rather than training separate networks for each quantile \citep{rodrigues2020beyond}. Note that the estimated quantiles share the network parameters except for the last layer.

\section{Experiments and results}
We evaluate our proposed approach on (i) A simulated dataset for density estimation; (ii) Variance estimation in the Fashion-MNIST dataset; and (iii) Lesion detection in a brain imaging dataset.  We compare our results qualitatively and quantitatively, using KL divergence between the learned distribution and the original distribution, for the simulated data with Comb-VAE \citep{skafte2019reliable} and VAE as baselines. We compare our lesion detection results with the VAE which estimates both the mean and the variance. The Area under the receiver operating characteristic curve (AUC) is used as a performance metric. 
We performed the lesion detection task by directly estimating a $95\%$ confidence interval. 
\begin{figure*}
\centering
  \includegraphics[width=1.0\textwidth]{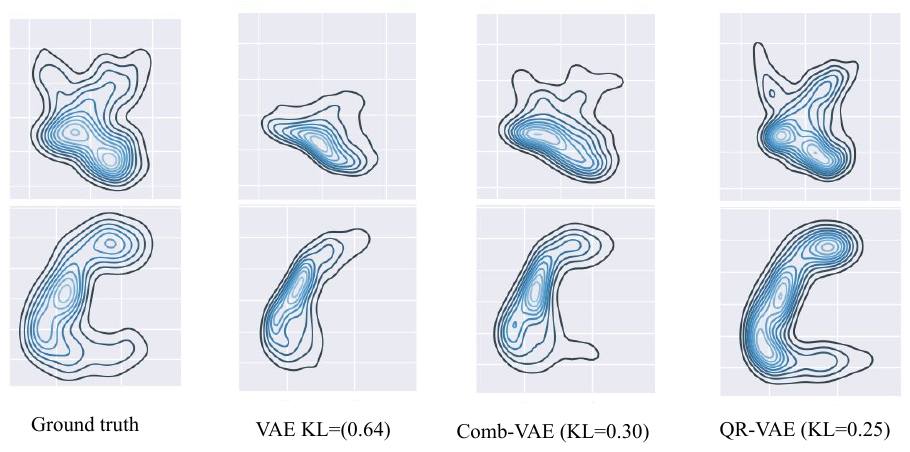}
  \caption{Pairwise joint distribution of the ground truth and distributions.
Top: $v_{1}$ vs. $v_{2}$ dimensions. Bottom: $v_{2}$ vs $v_{3}$ dimensions. From left to right: original distribution and distributions computed using VAE, Comb-VAE and QR-VAE, respectively. We also show the KL divergence between the learned distribution and the original distribution.}
  \label{fig:prob_dist}

\end{figure*}

\begin{figure*}[t]
\centering
  \includegraphics[width=1.0\textwidth]{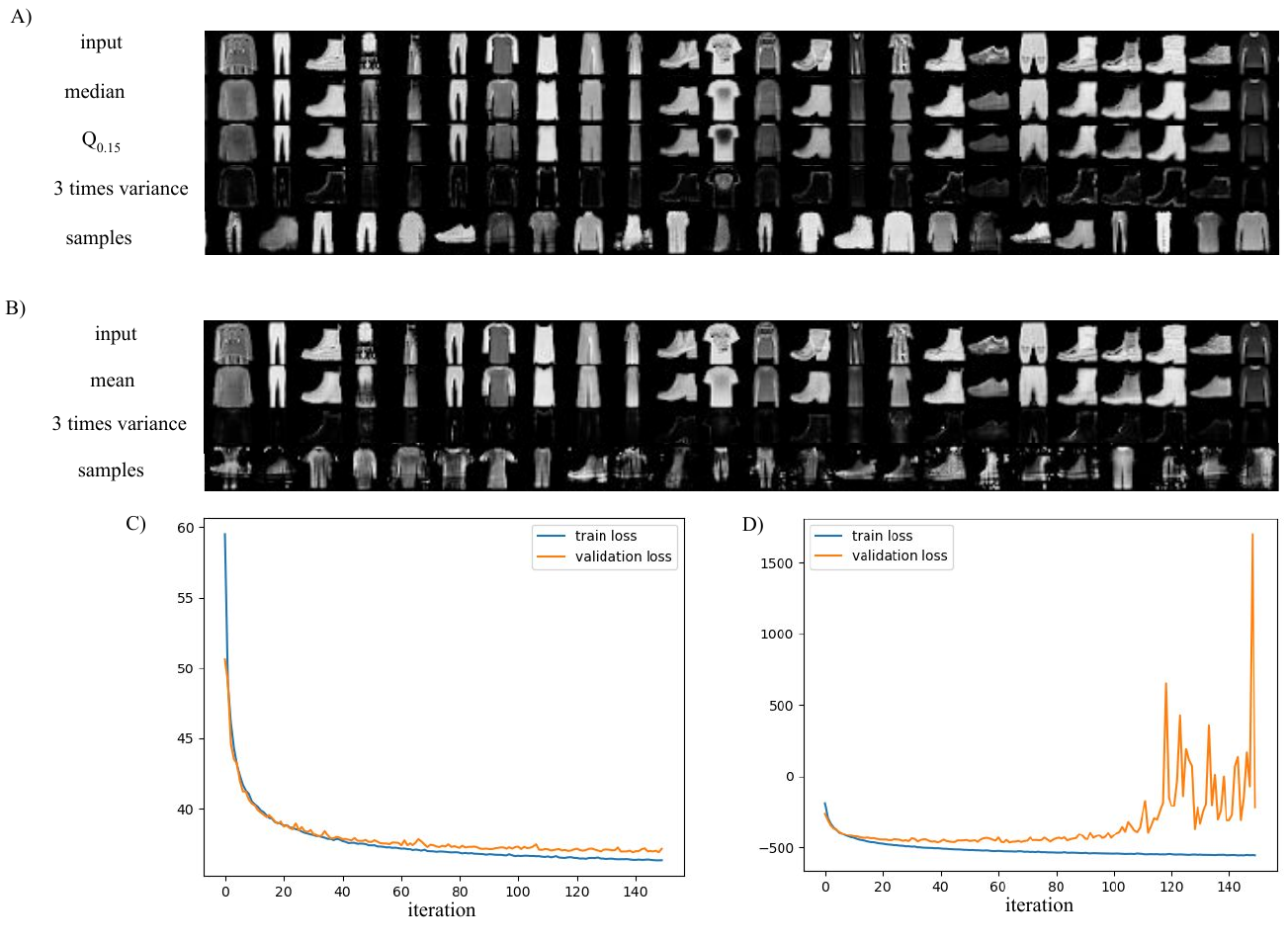}
  \caption{Estimating the variance in Fashion-MNIST dataset using A) QR-VAE where median ($Q_{0.5}$) and $Q_{0.15}$ quantiles are estimated. B) VAE where mean and variance are estimated. Training and validation loss curves for C) QR-VAE D) VAE.}
  \label{fig:anoma_det1}
\end{figure*}

\begin{figure*}[t]
\centering
  \includegraphics[width=1.0\textwidth]{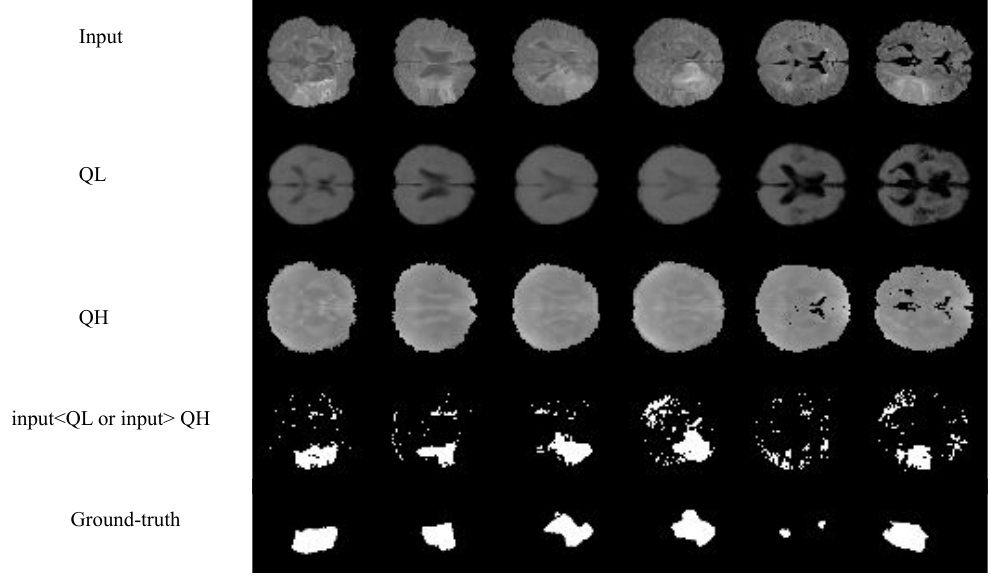}
  \caption{Lesion detection for ISLES dataset using $Q_L=Q_{0.025}$ and $Q_H=Q_{0.975}$. For lesion detection, we consider the pixels outside the [$Q_L$, $Q_H$] interval as outliers. Estimated quantiles are the outputs of QR-VAE.}
  \label{fig:isles_results2}
\end{figure*}

\subsection{Simulations}

Following \citet{skafte2019reliable}, we first evaluate our variance estimation using VAE, Comb-VAE, and QR-VAE on a simulated dataset. The two moon dataset inspires the data generation process for this dataset\footnote{\url{https://scikit-learn.org/stable/modules/generated/sklearn.datasets.make_moons}}. First, we generate 500 points in $\mathbb{R}^{2}$ in a two-moon manner to generate a known two-dimensional latent space. These are then are mapped to four dimensions  ($v_{1}$, $v_{2}$, $v_{3}$, $v_{4}$) by using the following equations:

\begin{align*}
v_{1}(z_{1},z_{2})=z_{1}-z_{2}+ \epsilon \sqrt {0.03+0.05(3+z_{1})}\\
v_{2}(z_{1},z_{2})=z_{1}^{2}-\frac{1}{2}z_{2}+ \epsilon \sqrt {0.03+0.03||{z_{1}}||_{2}}\\
v_{3}(z_{1},z_{2})=z_{1}z_{2}-z_{1}+ \epsilon \sqrt {0.03+0.05||{z_{1}}||_{2}}\\
v_{4}(z_{1},z_{2})=z_{1}+z_{2}+ \epsilon \sqrt {0.03+\frac{0.03}{0.02+||{z_{1}}||_{2}}}
\end{align*}

where $\epsilon$ is sampled from a normal distribution. For more details about the simulation, please refer to \citet{skafte2019reliable}\footnote{\url{https://github.com/SkafteNicki/john/blob/master/toy_vae.py}}. After training the models, we first sample from $z$ and then from the posteriors using the estimated means and variances from the decoder. The distribution of theses generated samples represents the learned distribution in the generative model. 

In Figure \ref{fig:prob_dist}, we plot pairwise joint distribution for the input data as well as the generated samples using various models. We used Gaussian kernel density estimation \citep{parzen1962estimation} for modeling the distributions from 1000 samples in each case. We observe that the standard VAE underestimates the variance resulting in insufficient learning of the data distribution. The samples from our QR-VAE model capture a data distribution more similar to the ground truth than either standard VAE or Comb-VAE. Our model also outperforms VAE and Comb-VAE in terms of KL divergence between input samples and generated samples as can be seen in Figure \ref{fig:prob_dist}. The KL divergence is calculated using universal-divergence, which estimates the KL divergence based on k-nearest-neighbor (k-NN) distance \citep{wang2009divergence}\footnote{\url{https://pypi.org/project/universal-divergence}}. 

\subsection{Estimating Variance in Fashion-MNIST}
In the second experiment, we trained VAE and QR-VAE on the Fashion-MNIST dataset \citep{xiao2017fashion}  where the outputs are: (i) mean and variance, and (ii) $0.15$, $0.5$ quantiles, respectively. The Fashion-MNIST dataset consists of 70,000 28x28 grayscale images of fashion products from 10 categories ($7,000$ images per category). 

The VAE and QR-VAE models for Fashion-MNIST experiments have the same architecture, except for the last layer. We use fully-connected layers with a single hidden layer consisting of 400 units both for the encoder and decoder. 
Figure \ref{fig:anoma_det1} shows the estimated variance along with the estimated mean for VAE and estimated quantiles for QR-VAE. We also show randomly generated samples for both models. It can be seen that the estimated variance using QR-VAE is structured and interpretable as it predicts a high variance around an object’s boundaries. The estimated variance using VAE is much lower than QR-VAE due to variance shrinkage. Moreover, instability in optimization of the VAE loss results in a diverging validation loss curve (Figure \ref{fig:anoma_det1}(D)), which indicates overfitting. The validation loss of QR-VAE is much better (Figure \ref{fig:anoma_det1}(E)). Furthermore, variance shrinkage in VAE
produces severely deteriorated sample quality due to mode collapse (Figure \ref{fig:anoma_det1}(D)), which is not the case for QR-VAE (Figure \ref{fig:anoma_det1}(C)). 

\subsection{Unsupervised Lesion Detection}
Finally, we demonstrate utility of the proposed QR-VAE for a medical imaging application of detecting brain lesions. 
Multiple automatic lesion detection approaches have been developed to assist clinicians in identifying and delineating lesions caused by congenital malformations, tumors, stroke or brain injury. The VAE is a popular framework among the class of unsupervised methods \citep{chen2018unsupervised,baur2018deep,pawlowski2018unsupervised}. After training a VAE on a lesion free dataset, presentation of a lesioned brain to the VAE will typically result in reconstruction of a lesion-free equivalent. The error between input and output images can therefore be used to detect and localize lesions. However, selecting an appropriate threshold that differentiates lesion from noise is a difficult task. Furthermore, using a single global threshold across the entire image will inevitably lead to a poor trade-off between true and false positive rates. 
Using the QR-VAE,  we can compute the conditional mean and variance of each output pixel. This allows a more reliable and statistically principled approach for detecting anomalies by thresholding based on computed p-values.

The network architectures of VAE, QR-VAE are chosen based on the previously established architectures \citep{larsen2015autoencoding}. Both the VAE and QR-VAE consist of three consecutive blocks of convolutional layer, a batch normalization
layer, a rectified linear unit (ReLU) activation
function and a fully-connected layer in the bottleneck
for the encoder. The decoder includes  three consecutive blocks
of deconvolutional layers, a batch normalization layer
and ReLU followed by the output layer that has 2 separate deconvolution
layers (for each output) with Sigmoid activations. For the VAE, the outputs represent mean and variance while for QR-VAE the outputs represent two quantiles from which the conditional mean and variance are computed at each voxel.
The size of the input layer is $3 \times 64 \times 64$ where the first dimension represents three different MRI contrasts:  T1-weighted, T2-weighted, and FLAIR for each image. 

For the  training, we use 20 central axial slices of brain MRI datasets 
from a combination of 119 subjects from the Maryland MagNeTS study 
\citep{gullapalli2011investigation} of neurotrauma and 112 subjects from the TrackTBI-Pilot 
\citep{yue2013transforming} dataset, both available for download from  
\url{https://fitbir.nih.gov}.
We use 2D slices rather than 3D images to make sure we had a large enough dataset for training the VAE. 
These datasets contain T1, T2, and FLAIR images for each
subject, and have sparse lesions. We have found that in practice both VAEs have some robustness to lesions in these training data so that they are sufficient for the network to learn to reconstruct lesion-free images as required for our anomaly detection task. The three imaging modalities  (T1, T2, FLAIR) were rigidly co-registered within subject and to the MNI brain atlas reference and re-sampled to 1mm isotropic resolution. Skull and other non-brain tissue were removed using BrainSuite (\url{https://brainsuite.org}). Subsequently, we reshaped each sample into $64 \times 64$ dimensional images and performed histogram equalization to a lesion free reference. We evaluated the performance of our model on a test set consisting of 20 central axial slices of 15 subjects from the ISLES (The Ischemic Stroke Lesion Segmentation) database \citep{maier2017isles} for which ground truth, in the form of manually-segmented lesions, is also provided. We performed similar pre-processing as for the training set.

\begin{figure*}
\centering
  \includegraphics[width=1.0\textwidth]{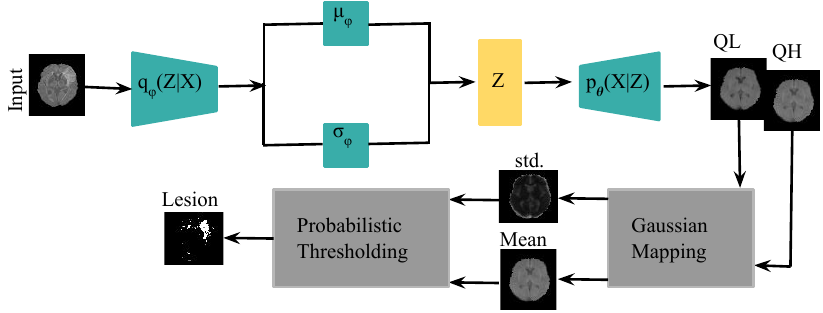}
  \caption{Estimating two quantiles in ISLES dataset using QR-VAE. Using the Gaussian assumption for the posterior, there is 1-1 mapping from these quantiles to mean and standard deviation.}
  \label{fig:anoma_det2}
\end{figure*}

For simplicity, we first performed the lesion detection task using the QR-VAE without the Gaussian assumption as shown in Figure \ref{fig:isles_results2}. We trained the QR-VAE to estimate the $Q_{0.025}$ and $Q_{0.975}$ quantiles. We then used these quantiles directly to threshold the input images for anomalies. This leads to a   5$\%$ false positive rate. This method is simple and avoids the need for validation data to determine an appropriate threshold. However, without access to p-values we are unable to determine a threshold that can be used to correct for multiple comparisons by controlling the false-disovery or family-wise error rate.

\begin{figure*}
\begin{center}
\includegraphics[height=1.25\textwidth]{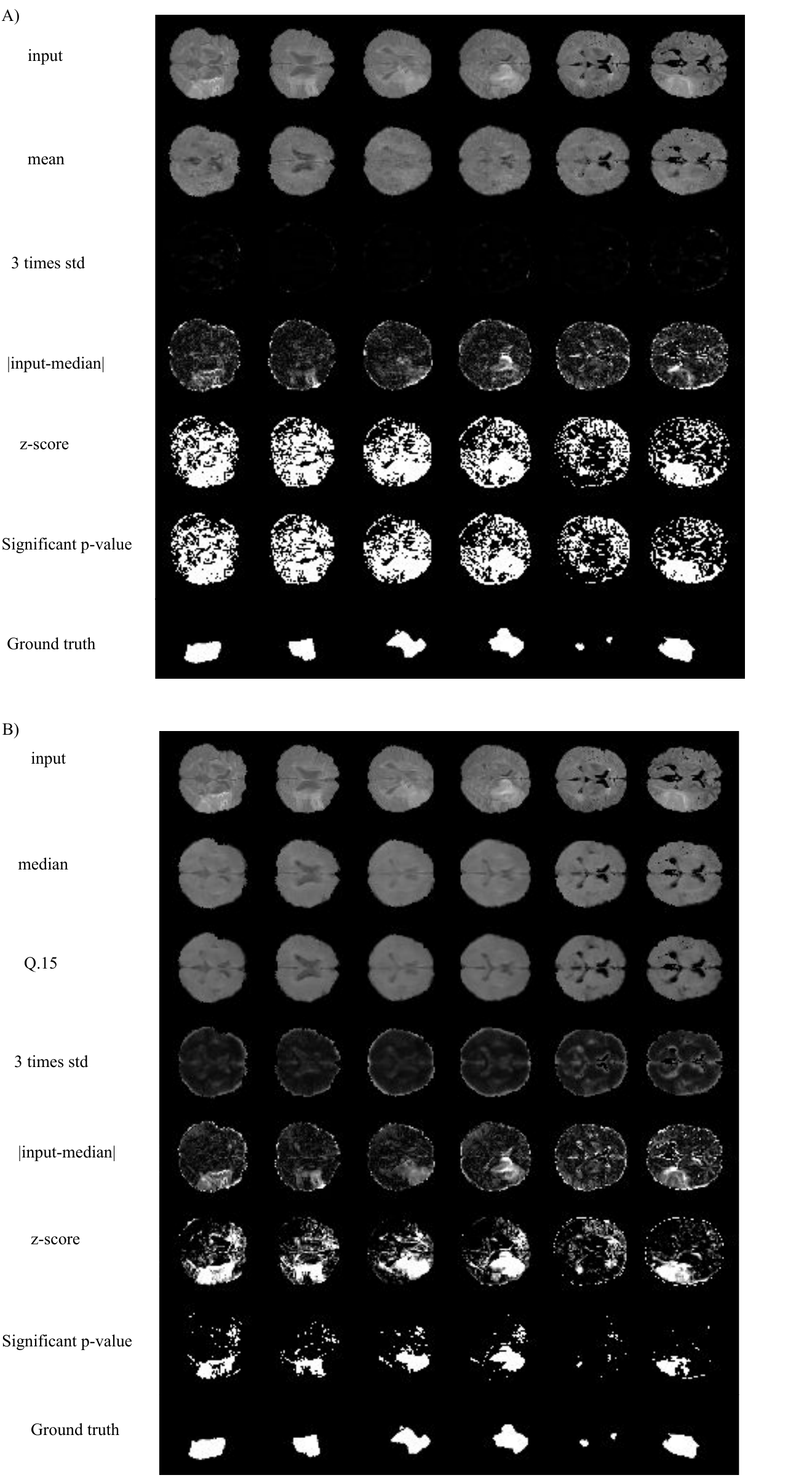}
\end{center}
\caption{Lesion detection for the ISLES dataset. A) VAE with mean and variance estimation  B) QR-VAE.
Firs, we normalize each pixel value using the pixel-wise model's estimates of mean and variance. The resulting z score is then converted to an FDR-corrected p-value and the images are thresholded at a significance level of 0.05. The bottom rows are ground truth based on expert manual segmentation of lesions.}
\label{fig:isles_results1}
\end{figure*}

In a second experiment, we trained a VAE with a Gaussian posterior and the QR-VAE as summarized in Figure \ref{fig:anoma_det2}, in both cases estimating conditional mean and variance. Specifically, we estimated the $Q_{0.15}$ and $Q_{0.5}$ quantiles for the QR-VAE and mapped these values to pixel-wise mean and variance. We then used these means and variances to convert image intensities to p-values. Since we are applying the threshold separately at each pixel, there is the potential for a large number of false positives simply because of the number of tests performed. For example, thresholding at an $\alpha =0.05$ significance level could result in up to 5$\%$ of the pixels being detected as anomalies. In practice the false positive rate is likely to be lower because of spatial correlations between pixels. To avoid an excessive number of false positives we threshold based on corrected p-values calculated to control the False Discovery Rate (FDR), that is the fraction of detected anomalies that are false positives \citep{benjamini1995controlling}. We chose the thresholds corresponding to an FDR corrected p-value of 0.05. As shown in Figure \ref{fig:isles_results1}, the VAE underestimates the variance, so that most of the brain shows significant p-values, even with FDR correction. On the other hand, the QR-VAE's thresholded results detect anomalies that reasonably match the ground truth. 
To produce a quantitative measure of performance, we also computed the area under the ROC curve (AUC) for VAE and QR-VAE. To do this we first computed z-score images by subtracting the mean and normalizing by standard deviation. We then applied a median filtering with a $7\times 7$ window. By varying the threshold on the resulting images and comparing it to ground truth, we obtained AUC values of 0.56 for the VAE and 0.94 for the QR-VAE.  

\section{Conclusion}
 Simultaneous estimation of the mean and the variance in VAE underestimates the true variance leading to instabilities in optimization \citep{skafte2019reliable}. For this reason, classical VAE formulations that include both mean and variance estimates are rarely used in practice. Typically, only the mean is estimated with variance assumed constant \citep{skafte2019reliable}. To address this problem in variance estimation, we propose an alternative quantile-regression model (QR-VAE) for improving the quality of variance estimation. We use quantile regression and leverage the Guassian assumption to obtain the mean and variance by estimating two quantiles. We show that our approach outperforms VAE as well as a Comb-VAE which is an alternative approach for addressing the same issue, in a synthetic dataset, the Fashion-MNIST dataset, and brain imaging datasets. Our approach also has more straightforward implementation compared to Comb-VAE. As a demonstrative application, we use our QR-VAE model to obtain a probabilistic heterogeneous threshold for a brain lesion detection task. This threshold results in a completely unsupervised lesion (or anomaly) detection method that avoids the need for a labeled validation dataset for principled selection of an appropriate threshold to control the false discovery rate.

\section*{Acknowledgements}       
 
This work is supported by the following grants: R01 NS074980, W81XWH-18-1-0614, R01 NS089212, and R01 EB026299. 

\bibliography{aistat_2021_conference}

\begin{thebibliography}{}

\bibitem[Aggarwal, 2015]{aggarwal2015outlier}
Aggarwal, C.~C. (2015).
\newblock Outlier analysis.
\newblock In {\em Data mining}, pages 237--263. Springer.

\bibitem[An and Cho, 2015]{an2015variational}
An, J. and Cho, S. (2015).
\newblock Variational autoencoder based anomaly detection using reconstruction
  probability.
\newblock {\em Special Lecture on IE}, 2(1):1--18.

\bibitem[Baur et~al., 2018]{baur2018deep}
Baur, C., Wiestler, B., Albarqouni, S., and Navab, N. (2018).
\newblock Deep autoencoding models for unsupervised anomaly segmentation in
  brain mr images.
\newblock In {\em International MICCAI Brainlesion Workshop}, pages 161--169.
  Springer.

\bibitem[Benjamini and Hochberg, 1995]{benjamini1995controlling}
Benjamini, Y. and Hochberg, Y. (1995).
\newblock Controlling the false discovery rate: a practical and powerful
  approach to multiple testing.
\newblock {\em Journal of the Royal statistical society: series B
  (Methodological)}, 57(1):289--300.

\bibitem[Blaauw and Bonada, 2016]{blaauw2016modeling}
Blaauw, M. and Bonada, J. (2016).
\newblock Modeling and transforming speech using variational autoencoders.
\newblock {\em Morgan N, editor. Interspeech 2016; 2016 Sep 8-12; San
  Francisco, CA.[place unknown]: ISCA; 2016. p. 1770-4.}

\bibitem[Chen and Konukoglu, 2018]{chen2018unsupervised}
Chen, X. and Konukoglu, E. (2018).
\newblock Unsupervised detection of lesions in brain {MRI} using constrained
  adversarial auto-encoders.
\newblock {\em arXiv preprint arXiv:1806.04972}.

\bibitem[Detlefsen et~al., 2019]{detlefsen2019reliable}
Detlefsen, N.~S., J{\o}rgensen, M., and Hauberg, S. (2019).
\newblock Reliable training and estimation of variance networks.
\newblock {\em arXiv preprint arXiv:1906.03260}.

\bibitem[Frazier, 2018]{frazier2018tutorial}
Frazier, P.~I. (2018).
\newblock A tutorial on bayesian optimization.
\newblock {\em arXiv preprint arXiv:1807.02811}.

\bibitem[Gullapalli, 2011]{gullapalli2011investigation}
Gullapalli, R.~P. (2011).
\newblock Investigation of prognostic ability of novel imaging markers for
  traumatic brain injury (tbi).
\newblock Technical report, BALTIMORE UNIV MD.

\bibitem[He, 1997]{he1997quantile}
He, X. (1997).
\newblock Quantile curves without crossing.
\newblock {\em The American Statistician}, 51(2):186--192.

\bibitem[Huang et~al., 2010]{huang2010active}
Huang, S.-J., Jin, R., and Zhou, Z.-H. (2010).
\newblock Active learning by querying informative and representative examples.
\newblock In {\em Advances in neural information processing systems}, pages
  892--900.

\bibitem[Kingma and Welling, 2013]{kingma2013auto}
Kingma, D.~P. and Welling, M. (2013).
\newblock Auto-encoding variational bayes.
\newblock {\em arXiv preprint arXiv:1312.6114}.

\bibitem[Koenker and Bassett~Jr, 1978]{koenker1978regression}
Koenker, R. and Bassett~Jr, G. (1978).
\newblock Regression quantiles.
\newblock {\em Econometrica: journal of the Econometric Society}, pages 33--50.

\bibitem[Koenker and Hallock, 2001]{koenker2001quantile}
Koenker, R. and Hallock, K.~F. (2001).
\newblock Quantile regression.
\newblock {\em Journal of economic perspectives}, 15(4):143--156.

\bibitem[Larsen et~al., 2015]{larsen2015autoencoding}
Larsen, A. B.~L., S{\o}nderby, S.~K., Larochelle, H., and Winther, O. (2015).
\newblock Autoencoding beyond pixels using a learned similarity metric.
\newblock {\em arXiv preprint arXiv:1512.09300}.

\bibitem[Maier et~al., 2017]{maier2017isles}
Maier, O., Menze, B.~H., von~der Gablentz, J., H{\"a}ni, L., Heinrich, M.~P.,
  Liebrand, M., Winzeck, S., Basit, A., Bentley, P., Chen, L., et~al. (2017).
\newblock {ISLES} 2015-a public evaluation benchmark for ischemic stroke lesion
  segmentation from multispectral {MRI}.
\newblock {\em Medical image analysis}, 35:250--269.

\bibitem[Makhzani et~al., 2015]{makhzani2015adversarial}
Makhzani, A., Shlens, J., Jaitly, N., Goodfellow, I., and Frey, B. (2015).
\newblock Adversarial autoencoders.
\newblock {\em arXiv preprint arXiv:1511.05644}.

\bibitem[Mattei and Frellsen, 2018]{mattei2018leveraging}
Mattei, P.-A. and Frellsen, J. (2018).
\newblock Leveraging the exact likelihood of deep latent variable models.
\newblock In {\em Advances in Neural Information Processing Systems}, pages
  3855--3866.

\bibitem[Parzen, 1962]{parzen1962estimation}
Parzen, E. (1962).
\newblock On estimation of a probability density function and mode.
\newblock {\em The annals of mathematical statistics}, 33(3):1065--1076.

\bibitem[Pawlowski et~al., 2018]{pawlowski2018unsupervised}
Pawlowski, N., Lee, M.~C., Rajchl, M., McDonagh, S., Ferrante, E., Kamnitsas,
  K., Cooke, S., Stevenson, S., Khetani, A., Newman, T., et~al. (2018).
\newblock Unsupervised lesion detection in brain ct using bayesian
  convolutional autoencoders.
\newblock {\em OpenReview}.

\bibitem[Reinhold et~al., 2020]{reinhold2020validating}
Reinhold, J.~C., He, Y., Han, S., Chen, Y., Gao, D., Lee, J., Prince, J.~L.,
  and Carass, A. (2020).
\newblock Validating uncertainty in medical image translation.
\newblock {\em arXiv preprint arXiv:2002.04639}.

\bibitem[Rodrigues and Pereira, 2020]{rodrigues2020beyond}
Rodrigues, F. and Pereira, F.~C. (2020).
\newblock Beyond expectation: deep joint mean and quantile regression for
  spatiotemporal problems.
\newblock {\em IEEE Transactions on Neural Networks and Learning Systems}.

\bibitem[Ross et~al., 2008]{ross2008bayesian}
Ross, S., Chaib-draa, B., and Pineau, J. (2008).
\newblock Bayesian reinforcement learning in continuous pomdps with application
  to robot navigation.
\newblock In {\em 2008 IEEE International Conference on Robotics and
  Automation}, pages 2845--2851. IEEE.

\bibitem[Skafte et~al., 2019]{skafte2019reliable}
Skafte, N., J{\o}rgensen, M., and Hauberg, S. (2019).
\newblock Reliable training and estimation of variance networks.
\newblock In {\em Advances in Neural Information Processing Systems}, pages
  6323--6333.

\bibitem[Stirn and Knowles, 2020]{stirn2020variational}
Stirn, A. and Knowles, D.~A. (2020).
\newblock Variational variance: Simple and reliable predictive variance
  parameterization.
\newblock {\em arXiv preprint arXiv:2006.04910}.

\bibitem[Szepesvari, 2010]{szepesvari2010algorithms}
Szepesvari, C. (2010).
\newblock Algorithms for reinforcement learning: Synthesis lectures on
  artificial intelligence and machine learning.
\newblock {\em Morgan and Claypool}.

\bibitem[Tagasovska and Lopez-Paz, 2019]{tagasovska2019single}
Tagasovska, N. and Lopez-Paz, D. (2019).
\newblock Single-model uncertainties for deep learning.
\newblock In {\em Advances in Neural Information Processing Systems}, pages
  6417--6428.

\bibitem[Volokitin et~al., 2020]{volokitin2020modelling}
Volokitin, A., Erdil, E., Karani, N., Tezcan, K.~C., Chen, X., Van~Gool, L.,
  and Konukoglu, E. (2020).
\newblock Modelling the distribution of 3d brain mri using a 2d slice vae.
\newblock In {\em International Conference on Medical Image Computing and
  Computer-Assisted Intervention}, pages 657--666. Springer.

\bibitem[Wang et~al., 2009]{wang2009divergence}
Wang, Q., Kulkarni, S.~R., and Verd{\'u}, S. (2009).
\newblock Divergence estimation for multidimensional densities via $ k
  $-nearest-neighbor distances.
\newblock {\em IEEE Transactions on Information Theory}, 55(5):2392--2405.

\bibitem[Wingate and Weber, 2013]{wingate2013automated}
Wingate, D. and Weber, T. (2013).
\newblock Automated variational inference in probabilistic programming.
\newblock {\em arXiv preprint arXiv:1301.1299}.

\bibitem[Xiao et~al., 2017]{xiao2017fashion}
Xiao, H., Rasul, K., and Vollgraf, R. (2017).
\newblock {Fashion-MNIST}: a novel image dataset for benchmarking machine
  learning algorithms.
\newblock {\em arXiv preprint arXiv:1708.07747}.

\bibitem[Yu and Moyeed, 2001]{yu2001bayesian}
Yu, K. and Moyeed, R.~A. (2001).
\newblock Bayesian quantile regression.
\newblock {\em Statistics \& Probability Letters}, 54(4):437--447.

\bibitem[Yue et~al., 2013]{yue2013transforming}
Yue, J.~K., Vassar, M.~J., Lingsma, H.~F., Cooper, S.~R., Okonkwo, D.~O.,
  Valadka, A.~B., Gordon, W.~A., Maas, A.~I., Mukherjee, P., Yuh, E.~L., et~al.
  (2013).
\newblock Transforming research and clinical knowledge in traumatic brain
  injury pilot: multicenter implementation of the common data elements for
  traumatic brain injury.
\newblock {\em Journal of neurotrauma}, 30(22):1831--1844.

\bibitem[Zhou et~al., 2020]{zhou2020estimating}
Zhou, T., Li, Y., Wu, Y., and Carlson, D. (2020).
\newblock Estimating uncertainty intervals from collaborating networks.
\newblock {\em arXiv preprint arXiv:2002.05212}.

\end{thebibliography}
\bibliographystyle{apalike}

\end{document}